# NPCs Vote! Changing Voter Reactions Over Time Using the Extreme AI Personality Engine

Jeffrey Georgeson, *Quantum Tiger Games, LLC*

*Abstract*—**Can non-player characters have human-realistic personalities, changing over time depending on input from those around them? And can they have different reactions and thoughts about different people? Using Extreme AI, a psychology-based personality engine using the Five Factor model of personality, I answer these questions by creating personalities for 100 voters and allowing them to react to two politicians to see if the NPC voters' choice of candidate develops in a realistic-seeming way, based on initial and changing personality facets and on their differing feelings toward the politicians (in this case, across liking, trusting, and feeling affiliated with the candidates). After 16 test runs, the voters did indeed change their attitudes and feelings toward the candidates in different and yet generally realistic ways, and even changed their attitudes about other issues based on what a candidate extolled.**

*Index Terms*—**artificial intelligence, personality, non-player characters, video games**

## I. Introduction and Objectives

Non-player characters are everywhere in most game worlds; they are the shopkeepers, the bureaucrats, the guards, the bartenders … in effect, almost all the inhabitants of any town or city. But they are as faceless as they are numerous, and they act in predetermined (and usually simplistic) ways that in no way allow them to act autonomously, and certainly not to react to events or change their minds over time in realistic ways.

This project focuses on creating NPCs with human-like personalities using the Extreme AI (ExAI) personality engine, and then testing whether or not these NPCs can react and change due to outside stimuli—in this case (and in this interesting US election year) to politicians each trying to win the votes of 100 NPCs. The NPCs will have personalities created using the ExAI engine, and these personalities will determine political leanings and whether a candidate is liked and/or trusted.

The primary objectives of this project are:

- To create NPCs with personalities utilizing the Extreme AI personality engine. These personalities should develop over time depending on interactions with the game environment, in this case two politicians. The personalities will be used to determine the characters' general political leanings (conservative vs liberal) and levels of trust and "liking" the political candidates.

- To stimulate the NPCs in such a way that they develop different attitudes toward each of the politicians, and then choose to vote for one or the other (or neither). With this, we show that the personality system can be used even for relatively quick changes such as attitudes toward specific characters (the politicians).[1] The attitudes should change in a realistic way—that is, they shouldn't be so jarringly strange that a player's immersion in the world of the sim would be lessened or destroyed.

Accomplishing the above objectives also satisfies a third, which is to show that ExAI is extensible to a number of complex situations in which NPC personalities can be used.

## II. What Is Extreme AI?

Extreme AI is a personality engine used to create evolving, human-like personalities for NPCs. It uses the Five Factor model of personality [1] to create 30 facet scores underlying the personality of each character (see Table 1). These scores range from 0-100, as they would for a real person rated on the NEO-PI (a test given to rate facets in the model; see [2]). Further, ExAI allows these facet scores to change over time, based on various models of personality change and elasticity [3, 4], although generally at a much faster rate than in real life so that a player can see these changes in game-time.[2]

In addition to allowing the developer to access the facets directly, ExAI ships with many multi-faceted response types (e.g., trust) that have been pre-created using tables of adjectives and their correlations to the Five Factor model from Costa and McCrae [5], John et al. [6], and Saucier and Ostendorf [7]; for example, trust can be construed as being highly correlated with the "trust" facet, and moderately correlated with three others—self-consciousness, altruism, and tender-mindedness. These pre-packaged types allow the developer to use more complex responses without having to research the interactions of facets and responses.

Although acting primarily on the underlying personality of the NPC, ExAI also can simulate more transient attitudes and feelings toward specific individuals; e.g., NPC Tilla Transit may feel kindness and trust toward a player (or character) who has been consistently nice to her and true to his word, but she will feel unkind and distrustful of another player who has been nothing but deceitful. She may still be a generally kind and trusting person overall (given her personality facets), but the influences of these two characters reinforce/override that. Note that these encounters also influence her base personality, but to a lesser extent (e.g., if she is surrounded by characters who are true to their word, she will become a more trusting person overall, increasing trust scores in her base personality).

---

[1] This project does not, however, do anything with the underlying personality changes that occur as NPCs interact with other characters over longer periods of time.

[2] Since time in-game is nearly always different from real-time, and normally personality changes would take place over months or years of real-time. Note that ExAI allows the developer to set the rate at which personalities change.

## III. PERSONALITY ENGINES AND EXTREME AI: RELATED WORK

Extreme AI addresses a problem still seen in many games: NPCs fail to act in a way that feels human, often acting as if they are actors with only one or two lines for every possible situation. This is the standard even for blockbusters such as Final Fantasy, and indeed is so standard that players expect to interact with NPCs only until the "loop" in speech begins to occur, then ignore that character forevermore. (And with good reason, as the NPC never develops any further, even if the world crumbles around him.)

While more extensive scripting can help maintain the illusion of NPC individuality a little longer, such scripting "tend[s] to constrain [NPCs] to a set of fixed behaviours which they cannot evolve in time with the world in which they dwell" [8], and these behaviors are "hard to extend, maintain and learn" [9]. (More recent research indicates that this is still the case; see [10].)

Several systems have attempted to give NPCs more flexibility in their interactions, typically either using personality or a social interaction system. For example, Mac Namee [11] uses Eysenck's 1965 [12] "two-dimensional classification" of personality, Lang's 1995 [13] "mood model", and a "relationship model" adding a "Level of Interest" value "indicating how interested one character is in another" to create an architecture to "drive the behaviours of *non-player support characters* in *character-centric* computer games" (italics in original) [11]. However, the limited number of personality traits do not change over time, no matter the stimulus, leading to a non-adaptive NPC.

Another personality system is Li & MacDonnell's [14] use of the Five Factor model (factors only) to create a base personality, with overlying social and emotion layers. In this system the social layer assigns the NPC membership in the social order, while the emotion layer is similar to MacNamee's mood layer and does all the changing. Again, the base personality doesn't change at all, and the emotions are transient, resulting in no long-term effects on the NPC.

More recently, Bura et al. [15] spoke of using the underlying facets of the Five Factor model in order to give NPCs personalities, which makes it more similar (at least in its use of model) to Extreme AI. It also uses combinations of

**TABLE 1. Facets of the Five Factor Model**

| ID | Facet Name | Big Five | Key/Notes—How Applies to NPC Traits |
|---|---|---|---|
| | | Openness | Openness to Experience: the active seeking and appreciation of experiences for their own sake |
| 1 | Fantasy | Openness | Receptivity to the inner world of imagination; low is solid, earthy reality; high lives a sort of fantasy life |
| 2 | Aesthetics | Openness | Appreciation of art and beauty; low finds this unimportant; high considers it above all else |
| 3 | Feelings | Openness | Openness to inner feelings and emotions; low is Mr. Spock, flat affect; high perceptive of others emotions, may be overinstructed by own emotions and feelings |
| 4 | Actions | Openness | Openness to new experiences on a practical level; high needs variety, novelty, change (travel, hobbies, etc.); low opposite |
| 5 | Ideas | Openness | Intellectual curiosity; low, not an idea person, not intellectually curious; high is full of ideas (may not act upon them), very interested in study |
| 6 | Values | Openness | Readiness to re-examine own values and those of authority figures; low individualises conservative; high tends to be liberal |
| | | Conscientiousness | Conscientiousness: degree of organization, persistence, control and motivation in goal directed behaviour |
| 7 | Competence | Conscientiousness | Belief in own self-efficacy; low indicates no self-belief; high believes self highly competent |
| 8 | Order | Conscientiousness | Personal organisation; low tends to be disorganised; high very orderly (bordering on obsessive) |
| 9 | Dutifulness | Conscientiousness | Emphasis placed on importance of fulfilling moral obligations; low not interested/worried about such obligations; high will try to fulfill them no matter the cost |
| 10 | Achievement Striving | Conscientiousness | Need for personal achievement and sense of direction; low doesn't care about achievements (grades, rewards, etc.); high has strong sense of purpose and high Aspiration levels (leadership, long-term plans, etc.) |
| 11 | Self-Discipline | Conscientiousness | Capacity to begin tasks and follow through to completion despite boredom or distractions; low no self-discipline (can't keep diet, etc.); high very disciplined |
| 12 | Deliberation | Conscientiousness | Tendency to think things through before acting or speaking; low jumps in without thinking; high deliberates, may be slow to act |
| | | Extraversion | Extraversion: quantity and intensity of energy directed outwards into the social world |
| 13 | Warmth | Extraversion | Interest in and friendliness towards others; low is uninterested, introverted; high is overly friendly |
| 14 | Gregariousness | Extraversion | Preference for the company of others; low prefers to be alone; high never wants to be alone |
| 15 | Assertiveness | Extraversion | Social ascendancy and forcefulness of expression; low holds back from expressing opinions; high forcefully talkative, opinionated |
| 16 | Activity | Extraversion | Pace of living; low is not energetic, lackadaisical; high full of energy and bounce |
| 17 | Excitement Seeking | Extraversion | Need for environmental stimulation; low not adventurous, no need for thrill-seeking; high loves risky activities, thrills for the sake of thrill |
| 18 | Positive Emotions | Extraversion | Tendency to experience positive emotions; low depressed or emotionally bland; high enthusiastic |
| | | Agreeableness | Agreeableness: the kinds of interactions an individual prefers from compassion to tough mindedness |
| 19 | Trust | Agreeableness | Belief in the sincerity and good intentions of others; low untrusting, unforgiving; high trusting, forgiving |
| 20 | Straightforwardness | Agreeableness | Frankness in expression; low guileful, deceitful, manipulative; high behaves sympathetically, undemanding, without guile, truthful |
| 21 | Altruism | Agreeableness | Active concern for the welfare of others; low critical, skeptical, not giving; high giving, warm, compassionate |
| 22 | Compliance | Agreeableness | Response to interpersonal conflict, willingness to defer to others during interpersonal conflict; low unforgiving, stubborn, uncooperative, offensive language, rebellious; High forgiving, cooperative, inoffensive, a pushover |
| 23 | Modesty | Agreeableness | Tendency to play down own achievements and be humble; low shows off, brags; high modest, humble |
| 24 | Tender-Mindedness | Agreeableness | Attitude of sympathy for others; low unsympathetic, not compassionate; high sympathetic, compassionate |
| | | Neuroticism | Neuroticism: identifies individuals who are prone to psychological distress |
| 25 | Anxiety | Neuroticism | Level of free floating anxiety; low calm, relaxed; high tense, anxious |
| 26 | Angry Hostility | Neuroticism | Tendency to experience anger and related states such as frustration and bitterness; low cheerful, not irritable, calm (in terms of irritability); high irritable, overreactive to frustration |
| 27 | Depression | Neuroticism | Tendency to experience feelings of guilt, sadness, despondency and loneliness; low optimistic, self-satisfied, happy; high guilt, discontent, pessimistic, feels cheated and victimised by life |
| 28 | Self-Consciousness | Neuroticism | Shyness or social anxiety; low has social poise, confidence; high shy, self-conscious, sensitive, easily embarrassed |
| 29 | Impulsiveness | Neuroticism | Tendency to act on cravings and urges rather than reining them in and delaying gratification; low clear-cut, consistent personality, not impulsive, can delay satisfaction; high impulsive, Moody, can't tolerate frustration, self-indulgent, inconsistent |
| 30 | Vulnerability | Neuroticism | General susceptibility to stress; low not prone to apprehension or stress, feel adequate; high stressed, gives up in the face of frustration, brittle ego-defenses, not self-confident |
| | | | Notes and applications from NEO PI-R manual reference online (Hogrefe, 20057); John et al, 2010; McCrae & Costa, 2010; Costa & McCrae, 1995 |

facets to create needs and behaviors (collectively called "traits"), which is similar to Extreme AI's response/stimulus types. However, NPCs are not given their own personality facets, but are instead "tagged" with the traits, which then can be compared with other tagged traits to create a scalar product that determines the character's course of action. For instance, a character tagged as Shy might be compared with a potential Seduction behavior, and the product of these would indicate that the NPC would never try (or would fail) at seduction. And, again, there is no indication that these traits would ever change in value (although you might be able to take the tag off the character).

Finally, Bura's model influenced the creation of the Love/Hate engine. Love/Hate [16] creates personalities for NPCs, but uses them as "faction templates" for entire groups of NPCs. Unlike Bura, Love/Hate gives each NPC his own faction template, which can use any of a number of personality systems, including the Five Factor model and its underlying facets. Love/Hate's focus, however, is on changes in relationships and emotions (using the Pleasure-Arousal-Dominance [17] model). These emotions do not affect the static underlying personality.

There are other personality/emotional models as well, but in general they are similar to those above: a changing set of emotions, combined with an unchanging underlying personality. The emotions do not carry over for extended periods of time (and it would be perhaps odd if they did), and in some cases are only generalized (that is, the NPC doesn't differentiate between the character with whom she's angry and the one with whom she's thrilled).

Social interaction engines are a newer and different kind of creature, and generally deal with social rules and the ways in which characters with different "traits" interact with these rules and social situations. For example, Versu [18, 19] gives each NPC desires and attributes, such as hating to be alone. These do not change; however, the characters evaluate those around them according to how well social roles and norms are being followed, and thus opinions (and relationships) can change over the course of the game. The attributes affect how well (or not well) an NPC will follow the norms and roles of the social situation around him.

Prom Week, which uses the Comme il Faut (CiF) social interaction engine, is similar to Versu and some others in that it represents social knowledge and rules to simulate interactions between characters [20]. While CiF uses rules to help characters navigate social exchanges, ExAI focuses primarily on the interior of the NPC—her personality—and builds from there. ExAI is not so much a social exchange engine as an individual personality engine, although of course it can help to model social situations (in much the same way as knowing the individual personalities involved at a party could lead to predictions about some of the social interactions that will occur). Also, the traits used in CiF are not personality traits, but items such as "attractive," "weakling," and "witty"—items generally describing outwardly perceived traits, not interior personality traits (which may or may not be able to be perceived from another's perspective); CiF's traits are more like some of the overlying response types in Extreme AI, which call upon combinations of underlying facets. However, even there the comparison is inexact; "attractive" would be in the eye of the beholder, including even the NPC herself, whose personality facets might lead her to believe she's unattractive when, in fact, others consider her beautiful. And, ultimately, these rules are imposed from without; the idea is to keep the characters operating within the social norms of the world, rather than to provide a high level of individuality.

Interestingly, Extreme AI could be used as an underlying element of social interaction systems, helping determine some of the traits used in the rules for social exchanges. It could also be used underneath the emotion systems, strengthening tendencies toward some emotions and weakening those toward others. It would provide a more realistic base personality in either kind of system.

No models were found that utilized personality in a truly human way—that is, that included sophisticated base personalities that would develop and change over time based on the NPCs' lived experiences, and would simultaneously allow for varied feelings toward individual others. ExAI, however, does exactly this. But how well does it work in the complex environment of a voting sim?

## IV.  USING EXTREME AI TO CREATE NPC VOTERS AND SETTING UP THE VOTING SIM

### A.  The Voters

For this simulation we create 100 voters of varying political and personal sensibilities and, given varying candidate scenarios, poll them as to who they would vote for (if they would vote at all). Ideally, over several runs:

- The voters would tend to follow their political and personal leanings, especially at first when the candidates are yet to be revealed; e.g., a staunch conservative would be extremely likely to vote conservative.
- The voters would take into account how they feel about the candidates (once revealed) and also how they feel about the candidates' actions (once those occur); to keep it simple, the voters are tested on how much they like each candidate and how much they trust each candidate. They are also tested on whether they would vote at all (given certain personality facets having to do with apathy).
- In each run with the same candidate actions, the overall effect should be similar (thus the actions of the entire population would make sense, given, for instance, a candidate doing good deeds and saying all the right things; said candidate could expect a boost overall in his or her polls).
- However, different individual NPCs might have slightly differing reactions in such a scenario, similar to the unpredictability of individual actions of real humans. While some of the initial personality setups would lend to a certain predictability (e.g., those who stand on the extremes of the political spectrum would rarely if ever find themselves voting for the opposing candidate), those closer to a middle stance would be more malleable (and thus be the targets of the politicians, just as in real life).

We begin with four evenly-divided overarching groups: Conservative, Liberal, and two sets of undecided voters (one group who is completely neutral [average in all personality], the other who is less likely to vote).

Within these, the Conservative and Liberal groups are subdivided into different personalities:

- 10 extreme (conservative or liberal): personality facets set up according to that which makes them have highly conservative or liberal tendencies (see Table 2 and explanation below)
- 10 fairly solid in their beliefs
- 5 leaning toward (conservative or liberal); most easily swayed

The "middle-voting" neutral group is constituted completely of those who have "average" personalities, the kind that are unlikely in real life but serve to be completely middle of the road politically.

The undecided group is politically middle-of-the-road, but is apathetic about the political process.

Note that, of course, the initial personality values change depending on what happens during the politics of the season. Note also that the program is set up to have much greater effects in a much shorter period of time than would be the case for a real election season or with real people, in order to get measurable reactions in a short scenario.

How did we determine which facet values to use? Using the same tables of adjectives and correlations as used in ExAI's other response types, conservatism was based on low scores in the facets for fantasy, aesthetics, ideas, and values. Note from the descriptions in Table 1 that a low score does not necessarily correlate to something negative; for instance, a low "values" facet score means only that the character does not like to re-examine her personal values nor those of authority figures [1, 2, 5, 6]. By extension, liberal NPCs were given high scores in these areas.

The facets used in the desire to vote were determined by the same tables of adjectives and combined scores in positive emotions and assertiveness, each with a high correlation.

### B. The Politicians

This is a campaign based on personalities and on one hot-button topic (rabbits overrunning the countryside). The voters choose a candidate and whether to vote based on how well-perceived a politician is—how well-liked, trustworthy, efficient, and dependable he is.

The politicians start out with a certain amount of baggage, presuming years of politicking and public opinion-forming. For instance, one combination is:

- Brian Jackson (Conservative): Very charismatic, a bit of a bumbler and often says awkward things; also doesn't always get things done.
- Len Kingston (Liberal): Not particularly likeable, and seen as a bit of a back-room dealer and hustler; he is, however, extremely efficient and much better spoken than Jackson.

The candidate's baggage can be reversed in the simulation, or they can each receive the same baggage. For this simulation, we begin with the same baggage for each candidate, then change this to see if the voters' opinions are changed.

The baggage is represented by making several calls to the ExAI engine, altering the attitudes of the voters toward these specific politicians (and more slightly changing their attitudes overall) using ExAI's AINoResult method:

AINoResult (politician, stimulus, posChange);

where politician is the name of the politician, stimulus is an attitude toward that politician, and posChange tells the engine whether the attitude is changed in a positive or negative way. For instance, to simulate the untrustworthiness of Kingston, the method would read:

AINoResult ("Kingston", "distrustful", true);

**TABLE 2. Facet Scores for Different Political Attitudes**

| | | |
|---|---|---|
| **very conservative:**<br>very low (10) fantasy, aesthetics, ideas, values<br>very high (80) dutifulness, trust<br>high (60) self-discipline, pos emo, assertiveness | **neutral:**<br>50 for each facet | **very liberal:**<br>very high (80) fantasy, aesthetics, ideas, values<br>very high (80) dutifulness, trust<br>high (60) self-discipline, pos emo, assertiveness |
| **conservative:**<br>very low (20) fantasy, aesthetics, ideas, values<br>high (60) dutifulness, trust<br>high (60) self-discipline, pos emo, assertiveness | **unwilling to vote:**<br>10 assertiveness, 10 pos emo<br>50 for all other facets | **liberal:**<br>high (70) fantasy, aesthetics, ideas, values<br>high (60) dutifulness, trust<br>high (60) self-discipline, pos emo, assertiveness |
| **leans conservative:**<br>low (30) fantasy, aesthetics, ideas, values<br>high (60) self-discipline, pos emo, assertiveness | | **leans liberal:**<br>high (60) fantasy, aesthetics, ideas, values<br>high (60) self-discipline, pos emo, assertiveness |

For this simulation, only four response types are used: distrust, kind, efficiency, and dependability. Distrust and kindness directly affect the attitudes of trusting and liking the candidate. Efficiency and dependability are compared to the voters' values for these and affect whether the voter feels like voting for a candidate (or for none at all).

### C. The Polls

Once the voters' attitudes have been adjusted for all the baggage of both candidates, we then check their current choice of candidate and change the color of the voter (represented by colored circles), as shown in Figure 1.

After the initial polls, the program allows for various actions by the candidates. For example, say we are following Brian Jackson in this simulation. In the first round, Jackson can:

- deliver a speech promising free public transport
- additionally, he'll lower taxes
- additionally, he'll upgrade the transport system

The first of these should give him a positive bump, at least among those who agree with him; this is represented by updating the voters' personalities slightly differently depending on their political affiliation. Conservatives increase their "kind" response through one AINoResult method call (thus liking Jackson a bit more). Liberals increase kindness, but also rise in distrust slightly (as they are more likely to be slightly suspicious of an opposition candidate). Those who are in the neutral and undecided categories like him a bit more, and don't change in the amount of trust they feel.

The second of Jackson's potential actions is, in effect, an intensification of the first. Conservatives have twice the increase in kindness (that is, the ExAI AINoResult method is called twice). Liberals have the same reaction as they have for the first of Jackson's options: increased kindness and distrust. Neutrals increase twice in kindness but also once in distrust. And finally, undecideds still increase only in kindness (just once).

The third option takes things over the top, and some voters (even in his own party) may find his promises preposterous. Conservatives and undecideds increase twice in kindness but also once in distrust. Liberals increase once in kindness and twice in distrust, as do neutrals. In the sim, we consistently choose this third option to see whether it really does have both a positive and negative effect.

In the second set of actions, a report is put out by a neutral group claiming Jackson's plans would cost too much. He can:

- ignore the report
- come out with his own, more positive report

The first is likely to damage his chances; the second may convince those who wish to believe in him, but will be looked on with suspicion by those who don't. The first option increases distrust in all voters, only once with conservatives and twice with all others. The second option causes those in

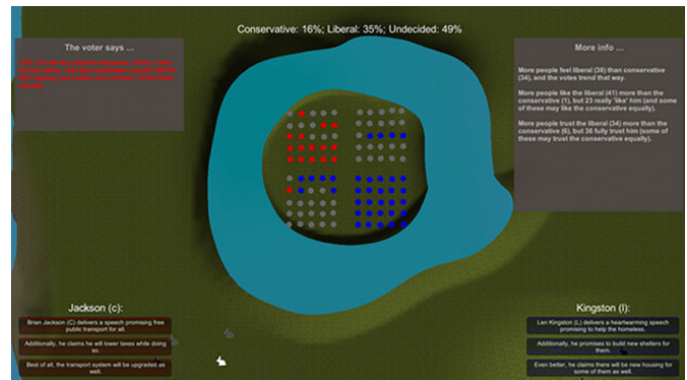

**Figure 1. Voting display and choices**

Jackson's party to have no reaction at all, and decreases the distrust felt by others by one call. In the simulation, we consistently choose the second option, trying to mitigate the damage done by the neutral group's report.

The player is allowed to make similar choices for Kingston instead, but his initial choice isn't as over-the-top, allowing us to see whether not over-promising helps the candidate more than promising too much (which it should, at least in this game world). His second choice is the same as Jackson's.

Note that, although the voters receive the same stimuli (method calls based on the candidates' choices), individual reactions can vary slightly; this is built into the ExAI engine, using fuzzy logic for NPCs' reactions. This variation is meant to allow for individualism and differences in day-to-day character reactions to stimuli (say, if the NPC is having a bad day or a great one) without creating odd reactions that make no sense and take away from game immersion (as would happen with a purely random reaction). In this simulation, we obtain the voters' reactions through polling (in the political sense).

Overall, there are four polling points in the first part of the simulation:

- before any candidates are revealed: to show how the voters would vote with only political party affiliation
- after candidates' personalities (and attendant baggage) are revealed: operates on the presumption that the voters have known these candidates for a long time and are judging them on past actions
- after the politician's first action
- after the politician's second action

This is followed by the introduction of a specific issue: rabbits are overrunning the countryside! In addition to how the voters feel about the politicians, they are now polled about the rabbits. How the politicians react to this issue determines the results of the next three polls, culminating in the final tally: in general, Jackson can:

- Ignore the rabbits
- Make a joke about them
- Show he's tough on rabbits
- Use the rabbits negatively against his opponent

Kingston can:

- Ignore the rabbits
- Do something showing he loves the rabbits
- Show he's tough on rabbits
- Use the rabbits negatively against his opponent

These have slightly different effects depending on a voter's feelings about the rabbits, in addition to party affiliation. For instance, if Jackson shows he's tough on rabbits and tries to get rid of them, those in his own party who like rabbits will decrease in their "like" of Jackson. Those who don't like rabbits, however, will increase in their like of Jackson. Similar reactions occur with the other voting blocs and for Kingston's choices.

Why so many polls? To get a better sense of how voters' attitudes are adjusting over as many stimuli as possible, and so any intermediate outliers will be more likely to be seen.

## V. TESTING & RESULTS: THE VOTERS LISTEN, AND VOTE

### A. Test runs

Multiple runs with different conditions were required to gauge the voters' reactions to various political stimuli and to see whether these reactions seemed believable. Ten runs were performed with the candidates having the same baggage (both likeable). For the first five runs we chose from Jackson's point of view, going with choice three (overpromising) in the first round and choice two (providing his own report) in the second. After this, we chose Jackson joking about rabbits, then claiming he'll get rid of them, and then talking about building a fence. For consistency's sake, these choices were made regardless of the voters' feelings about rabbits.

The next five runs were from Kingston's viewpoint. We chose the second choice in the first round (not overpromising), then choice two (providing his own report). For the rabbits, Kingston is very positive, first loving them, then kissing them (to show they're lovable), and then waffling and talking about building a fence. Again, the rabbit choices were made regardless of the public's preference for or against rabbits.

After this, we changed the baggage so that Jackson was likeable and Kingston was not (as in the example at the beginning of section IVb). After three more runs using the same answers for Jackson, we reversed these and ran it again the same number of times.

### B. Results

Overall, regardless of the candidates themselves, the political leanings of the voters remain about the same (around 25 each party, with neutrals and undecideds still neutral and undecided; see Figure 2); the real changes are in whether the candidates are liked and/or trusted, especially vs one another.

Initial likes are low (Figure 3), which makes sense: only a few voters (the most staunch of their parties) automatically like the candidate sight-unseen. Only once do these initial likes reach double digits (12). The exact number of voters "liking" in the initial poll varies, however, which mirrors the never-exact way in which opinions work in the real world.

Initial trust (Figure 4) averages slightly higher (high single digits to low double digits), which again makes sense: Voters identifying with a political party will tend to trust an unknown candidate within that party, generally speaking, more than a candidate in the opposing party—if they trust anyone more at all. Again, it is the party stalwarts who tend to have this view more than others.

Once the candidates appear, the neutral and undecided voters' opinions can fluctuate quite a bit, at least at first, and some choose a candidate, but generally only a handful (Figure 5). When the candidates have the same baggage (in this case, being likeable, etc.), they tend to lose very few votes (because they tend not to lose votes from party members), but can gain as many as ten; in some cases, both candidates gain votes.

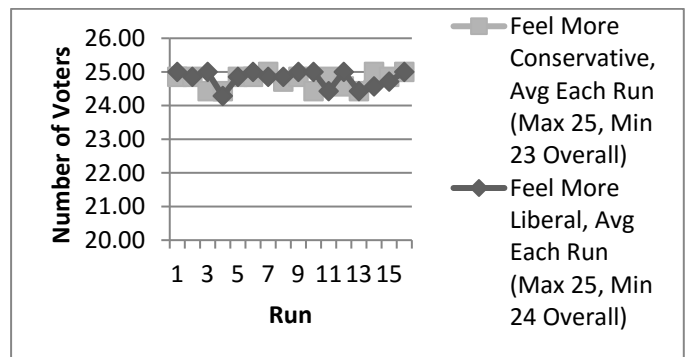

Figure 2. Political Leanings of Voters, Average Each Run

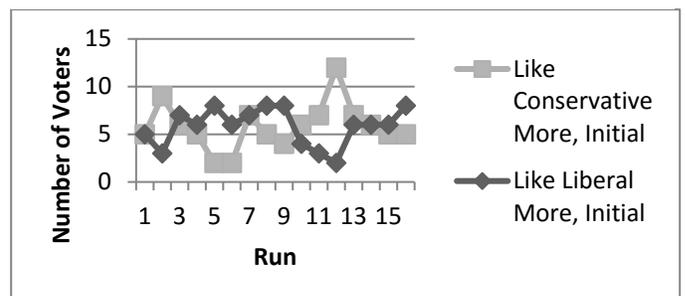

Figure 3. Voters Initially Liking One Candidate More

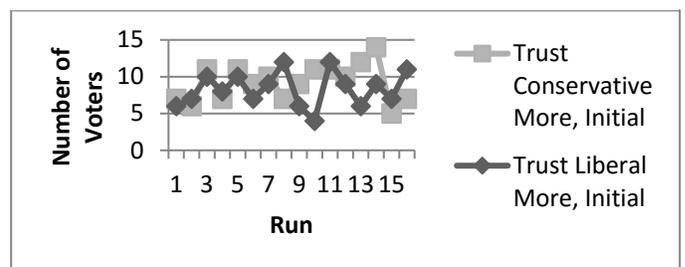

Figure 4. Voters Initially Trusting One Candidate More

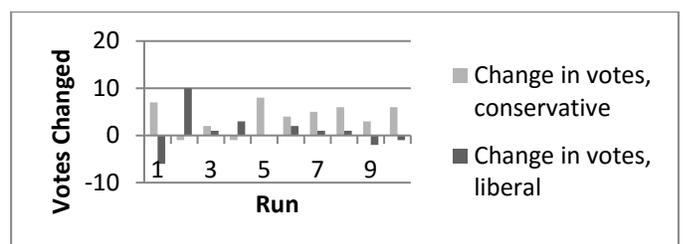

Figure 5. Votes Changed After Candidate Personalities Chosen

Once candidates begin their own attempts at manipulating the vote, things become more complicated. Jackson's answer (that goes over the top) does not have a consistent effect: More often than not he gains a few votes, but so does Kingston (Figure 6). Kingston's promises in the first round fare better, with that candidate almost always gaining votes and Jackson losing votes more often than not (Figure 7).

Neither candidate's second answer (refuting the independent report) is very convincing to voters, as both more often than not lose a few votes (Figures 8 and 9). However, Jackson seems to do better overall, perhaps indicating that the conservative voters are slightly more willing to stand behind their candidate than are the liberals—although this is not borne out in either the trust or like levels reported by the voters, which don't correlate well with the changes in votes at this stage (going up while the votes for a candidate go down or vice versa about half the time, but going up as the candidate's votes go up or vice versa the other half).

Jackson's answers to the rabbit questions are made without regard to the way the public feels about the rabbits, but this doesn't always correlate with his outcomes (see Table 3). Joking about the rabbits doesn't seem to help him or hurt him much; he loses an average of a little more than one vote in this round, which makes sense, as the public don't see his joke as really staking out a position. Saying he'll get rid of the rabbits in the second round shouldn't help him much, as the public is more evenly divided on the rabbits, but for some reason he gains an average of one vote (and the gains/losses in each run don't correlate to the number of voters disliking/liking rabbits, although the numbers aren't large). His answer about fences does not help him (he averages no gain or loss), and the voters are again nearly evenly divided on the rabbits; this fits better with what we'd expect.

Kingston's embrace of the rabbits tends to cause him problems when more voters dislike the rabbits (by far the case during the first rabbit action), and tends to help him slightly when more voters like the rabbits (which is the case for the second rabbit action; see Table 4). When Kingston waffles and talks about fences as his third action, the voters (again leaning toward liking rabbits) give him no love. Interestingly, his embrace of rabbits seems to affect the voters' perception; significantly more voters like the rabbits by the end of the polls than at the beginning. This wasn't an intended outcome, but does seem to be realistic—when a public figure comes out strongly in favor of something, people listen (although the reaction in the sim may be too strong).

When we change the initial baggage to be different for each candidate, it gives a very definite boost to the candidate who is more liked and trusted, usually a double-digit boost. After this the reactions are similar to those in previous rounds, which means that the trailing candidate never manages to catch up and indicates that prior baggage in one's political career is extremely important, which again would seem to be true of the real world as well.

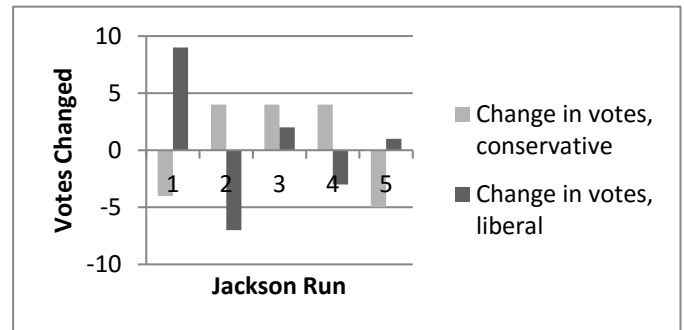

**Figure 6. Change in Votes After Jackson's OTT Answer**

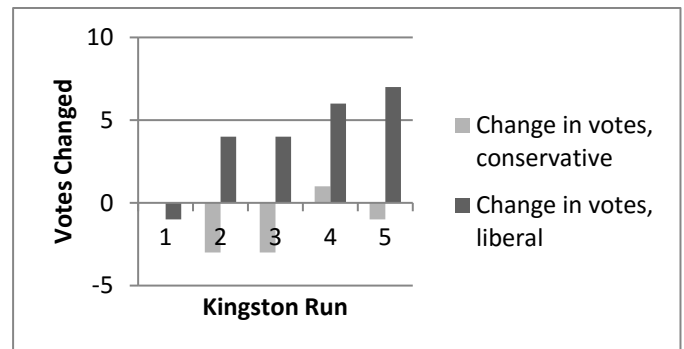

**Figure 7. Change in Votes After Kingston's Non-OTT Answer**

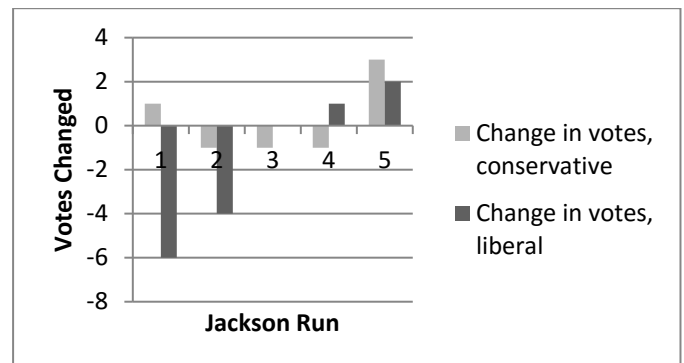

**Figure 8. Change in Votes After Jackson Refutes Report**

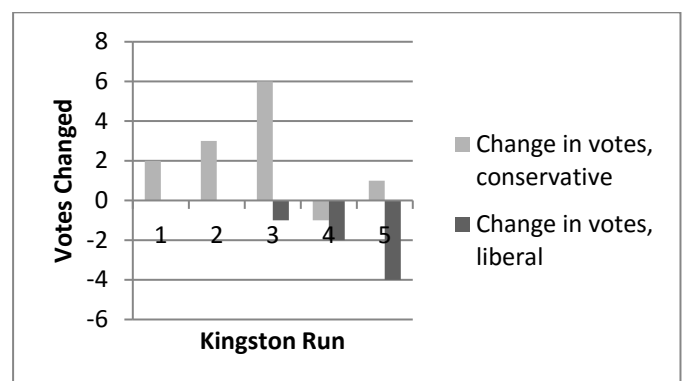

**Figure 9. Change in Votes after Kingston Refutes Report**

| Table 3. Jackson Rabbit Runs | | Table 4. Kingston Rabbit Runs | |
| --- | --- | --- | --- |
| Net Like Rabbits | Change in Votes | Net Like Rabbits | Change in Votes |
| Jokes about Rabbits, Five Runs | | Loves Rabbits, Five Runs | |
| -13 | -1 | -44 | -4 |
| -9 | -2 | -43 | -4 |
| -9 | -3 | -42 | -2 |
| -10 | 1 | -39 | 0 |
| -12 | 0 | -42 | -8 |
| Gets Rid of Rabbits, Five Runs | | Really Loves Rabbits, Five Runs | |
| -4 | 3 | 2 | 1 |
| -1 | 1 | 1 | 1 |
| -1 | 2 | 2 | 1 |
| -3 | -2 | 5 | -1 |
| -4 | 1 | 8 | 6 |
| Build a Fence | | Build a Fence | |
| 0 | -2 | 1 | -2 |
| 2 | -1 | 2 | 0 |
| -2 | 2 | 3 | 0 |
| 1 | 3 | 5 | 0 |
| -2 | -2 | 8 | 0 |

## VI. Conclusions

As with previous trials of the ExAI engine [21, 22], this simulation demonstrates the ability of NPCs to change and adapt due to outside stimuli (in this case politicians) in a realistic manner overall (with a few minor exceptions as noted above); or, as defining "realistic" is rather subjective, at the very least this voting sim shows that the voters' reactions are not unrealistic—that is, they do not take the player out of the game by acting in counterintuitive ways, and in fact maintain interest and immersion by reacting in individual, yet generally realistic, ways—including changing their opinions based on the strong conviction of a public figure, which wasn't an objective but occurred anyway.

This project is admittedly rather simplistic in terms of re-creating the motivations of voters in the real world, and necessarily arbitrary (as the developer defines the voters' personality changes due to the politicians' actions). A different implementation might define the voters' reactions even more generically (not changing the method calls depending on party affiliation) and allow the voters to make decisions strictly based on their internal personality structure. Further, the ExAI engine can be used to make the voters' reactions as complex as desired by the developer, possibly including not only liking and trusting candidates, but using a host of underlying desires as well (say, some NPCs favor certain specific issues, or there are relationships and intrigue that underlie some of the candidates' actions of which some voters are aware, etc.). Also, a deeper study of voting preferences vis à vis personality would likely yield a more thorough understanding of what could be added or changed in the way the voting sim is set up, creating even more realistic patterns that could be used not only as a game, but perhaps even as a predictive model for actual elections.

Finally, one could take advantage of the changing of personalities over time in ExAI to create an extended simulation over several elections; NPCs could thus change their political affiliations over time, as can occur in real life.

## Acknowledgments

Thanks to Dr. Christopher Child, who was the thesis advisor for the original paper describing what became Extreme AI, and who helped edit this paper.